\documentclass[sigconf,camera-ready]{acmart}
\AtBeginDocument{%
  \providecommand\BibTeX{{%
    \normalfont B\kern-0.5em{\scshape i\kern-0.25em b}\kern-0.8em\TeX}}}

\copyrightyear{2022} 
\acmYear{2022} 
\setcopyright{rightsretained} 

\acmConference[MM '22]{Proceedings of the 30th ACM International Conference on Multimedia}{October 10--14, 2022}{Lisboa, Portugal}
\acmBooktitle{Proceedings of the 30th ACM International Conference on Multimedia (MM '22), Oct. 10--14, 2022, Lisboa, Portugal}
\acmISBN{978-1-4503-9203-7/22/10}
\acmDOI{10.1145/3503161.3547906}

\usepackage{etoolbox}
\makeatletter
\patchcmd{\maketitle}{\@copyrightpermission}{
   \begin{minipage}{0.3\columnwidth}
     \href{https://creativecommons.org/licenses/by-nc-sa/4.0/}{\includegraphics[width=0.90\textwidth]{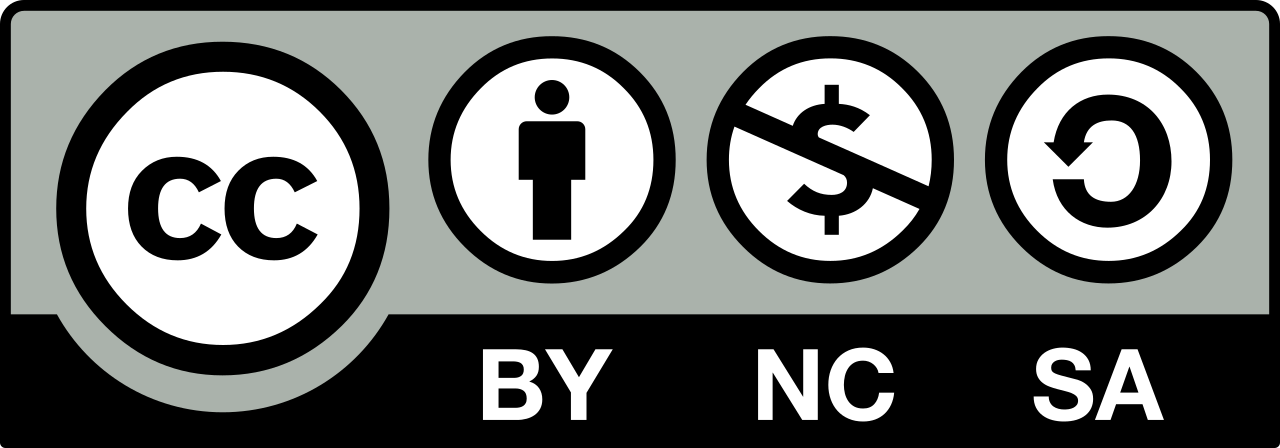}}
   \end{minipage}\hfill
   \begin{minipage}{0.7\columnwidth}
     \href{https://creativecommons.org/licenses/by-nc-sa/4.0/}{This work is licensed under a Creative Commons Attribution-NonCommercial-ShareAlike International 4.0 License.}
   \end{minipage}
  
   \vspace{5pt}
}{}{}

\makeatother

\usepackage{microtype}
\usepackage{amsmath}
\usepackage{graphicx}
\usepackage{subfigure}
\usepackage{booktabs}
\usepackage{multirow}
\usepackage{amsmath}
\usepackage{dsfont}
\usepackage{bbding}
\usepackage{pifont}
\usepackage{balance}

\begin{document}

%%%%%%%%%%%Templates%%%%%%%%%%%%
% https://www.acm.org/publications/proceedings-template

\title{MIntRec: A New Dataset for Multimodal Intent Recognition}

\author{Hanlei Zhang}
\affiliation{
\institution{State Key Laboratory of Intelligent Technology and Systems, Department of Computer Science and Technology}
\country{Tsinghua University, Beijing, China}
}
\email{zhang-hl20@mails.tsinghua.edu.cn}
\author{Hua Xu*}
\affiliation{
    \institution{State Key Laboratory of Intelligent Technology and Systems, Department of Computer Science and Technology, }
    \country{Tsinghua University, Beijing, China}
}
\email{xuhua@tsinghua.edu.cn}
\author{Xin Wang}
\affiliation{
    \institution{Department of Computer Science and Technology, Tsinghua University;}
    \country{
    School of Information Science and Engineering, Hebei University of Science and Technology}
}
\email{wx\_hebust@163.com}
\author{Qianrui Zhou}
\affiliation{
   \institution{State Key Laboratory of Intelligent Technology and Systems, Department of Computer Science and Technology, }
    \country{Tsinghua University, Beijing, China}
}
\email{zhougr18@mails.tsinghua.edu.cn}
\author{Shaojie Zhao}
\affiliation{
    \institution{Department of Computer Science and Technology, Tsinghua University;} 
    \country{School of Information Science and Engineering, Hebei University of Science and Technology}
}
\email{murrayzhao@163.com}
\author{Jiayan Teng}
\affiliation{
    State Key Laboratory of Intelligent Technology and Systems, Department of Computer Science and Technology, \country{Tsinghua University, Beijing, China}
}
\email{tengjy20@mails.tsinghua.edu.cn}

\makeatletter
\def\authornotetext#1{
\if@ACM@anonymous\else
    \g@addto@macro\@authornotes{
    \stepcounter{footnote}\footnotetext{#1}}
\fi}
\makeatother
\authornotetext{Corresponding author.}

\renewcommand{\shortauthors}{Hanlei Zhang et al.}
\renewcommand{\authors}{Hanlei Zhang, Hua Xu, Xin Wang, Qianrui Zhou, Shaojie Zhao, Jiayan Teng}
%%%%%%%%%%%%%%%%%%%%%%%%%%%%%%%%%%%%%%
% \author{
% Hanlei Zhang\textsuperscript{\rm 1,2}, 
% Hua Xu\textsuperscript{\rm 1,2}*,
% Xin Wang\textsuperscript{\rm 1,3},
% Qianrui Zhou\textsuperscript{\rm 1,2},
% Shaojie Zhao\textsuperscript{\rm 1,3},
% Jiayan Teng\textsuperscript{\rm 1,2}
% }

% \makeatletter
% \def\authornotetext#1{
% \if@ACM@anonymous\else
%     \g@addto@macro\@authornotes{
%     \stepcounter{footnote}\footnotetext{#1}}
% \fi}
% \makeatother
% \authornotetext{Corresponding author.}

% \renewcommand{\shortauthors}{Hanlei Zhang et al.}
% \renewcommand{\authors}{Hanlei Zhang, Hua Xu, Xin Wang, Qianrui Zhou, Shaojie Zhao, Jiayan Teng}

% \affiliation{
% \textsuperscript{\rm 1}State Key Laboratory of Intelligent Technology and Systems, Department of Computer Science and Technology,\\Tsinghua University, Beijing 100084, China;
%  \textsuperscript{\rm 2}Beijing National Research Center for Information Science and Technology (BNRist), Beijing 100084, China; \textsuperscript{\rm 3}School of Information Science and Engineering,  \\
% \text{
%     zhang-hl20@mails.tsinghua.edu.cn, xuhua@tsinghua.edu.cn, wx\_hebust@163.com
% }\\
% \text{
%     zhougr18@mails.tsinghua.edu.cn, 
%     tengjy20@mails.tsinghua.edu.cn, murrayzhao@163.com
% }
%%%%%%%%%%%%%%%%%%%%%%%%%%%%%%%%%%%%%%

\begin{abstract}
Multimodal intent recognition is a significant task for understanding human language in real-world multimodal scenes. Most existing intent recognition methods have limitations in leveraging the multimodal information due to the restrictions of the benchmark datasets with only text information. This paper introduces a novel dataset for multimodal intent recognition (MIntRec) to address this issue. It formulates coarse-grained and fine-grained intent taxonomies based on the data collected from the TV series Superstore. The dataset consists of 2,224 high-quality samples with text, video, and audio modalities and has multimodal annotations among twenty intent categories. Furthermore, we provide annotated bounding boxes of speakers in each video segment and achieve an automatic process for speaker annotation. MIntRec is helpful for researchers to mine relationships between different modalities to enhance the capability of intent recognition. We extract features from each modality and model cross-modal interactions by adapting three powerful multimodal fusion methods to build baselines. Extensive experiments show that employing the non-verbal modalities achieves substantial improvements compared with the text-only modality, demonstrating the effectiveness of using multimodal information for intent recognition. The gap between the best-performing methods and humans indicates the challenge and importance of this task for the community. The full dataset and codes are
available for use at \url{https://github.com/thuiar/MIntRec}.
  \end{abstract}




\maketitle

\section{Introduction}
\begin{figure}[t!]\small
	\centering  
	\includegraphics[scale=.35]{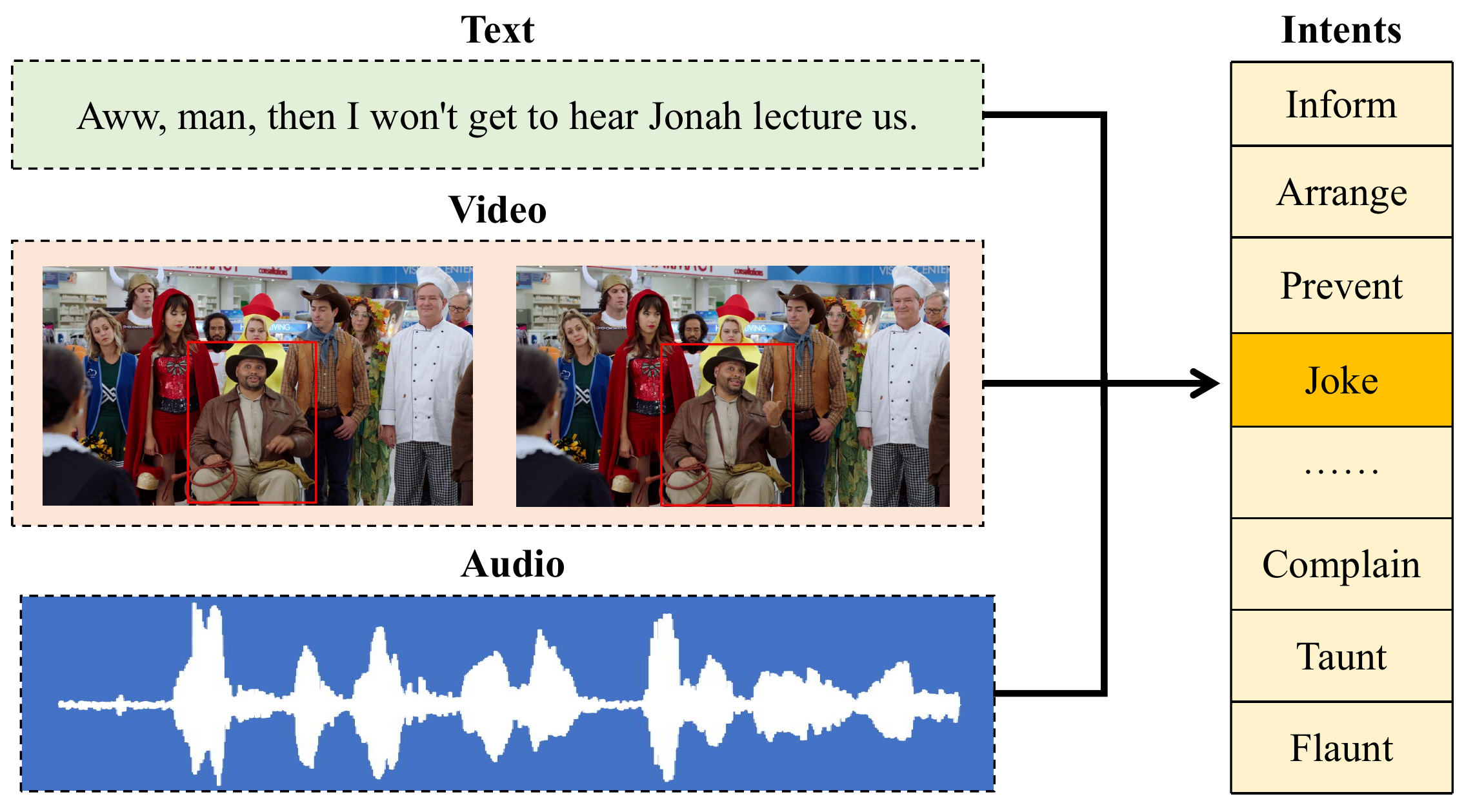}

% However, most intent recognition benchmark datasets contain only text information, leading to
	\caption{\label{example} An example of multimodal intent recognition.}
\end{figure}
\begin{figure*}[t!]\small
	\centering  
	\includegraphics[scale=.5]{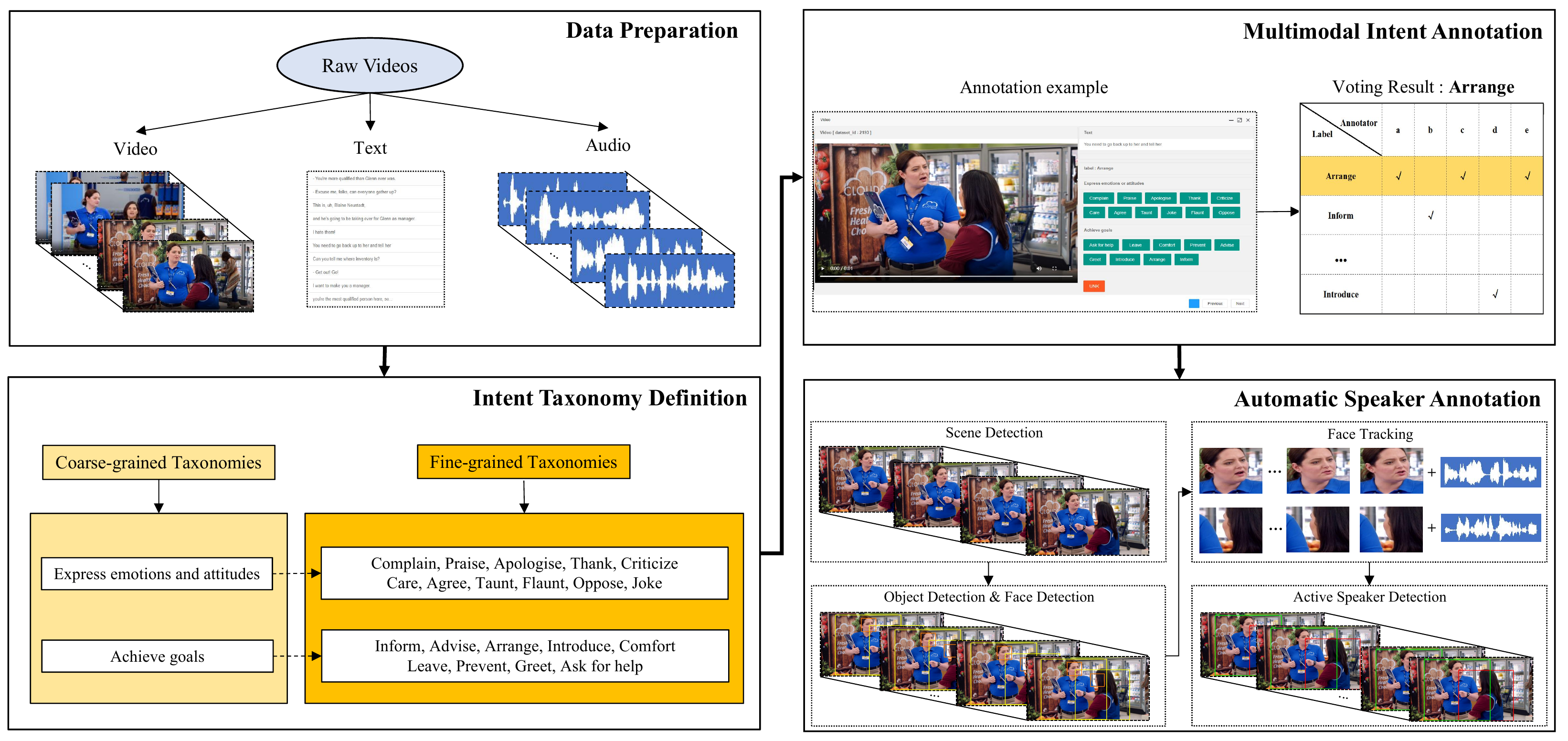}
	\caption{\label{pipeline} The process of building the  MIntRec dataset.
}
\end{figure*}
Intent recognition is crucial in natural language understanding (NLU), which aims to leverage the text information to determine the intent categories for better conversational interactions. Though text-based intent recognition has achieved remarkable performance~\cite{chen2019bert, qin-etal-2019-stack,zhang-etal-2021-textoir}, it mainly focuses on goal-oriented tasks in specific domains. The intents of these tasks usually come from orders or queries with clear semantic features~\cite{coucke2018snips,casanueva2020efficient}, which are different from the real-world multimodal language with rich emotional, attitudinal and behavioral information. Combining natural language with non-verbal signals (e.g., expressions, body movements, and tone of speech) may be beneficial in analyzing human intentions 
from multiple perspectives and provide more friendly services. 

% which may be inapplicable in real-world conversational scenes. 

% , which may be different from 

% it lacks the utilization of the knowledge from non-verbal modalities, which may be inapplicable in real-world multimodal scenes. 

% Therefore, we focus on multimodal intent recognition in this work. 

%   which may be rare in the real-world scenes.   combine language with other non-verbal behaviors (e.g., expressions, body movements, and tone of speech) to identify a person's real purpose. Multimodal information is 

% However, in real-world multimodal scenes, many non-verbal behaviors  may also be beneficial to intent analysis. 

Taking Figure~\ref{example} as an example, we might infer the speaker to be complaining about someone based on the text information. After combining the video and audio information, we find the real intention is joking, as the speaker's expression and tone are cheerful rather than indignant. It 
indicates that using text alone has difficulties satisfying the requirements of identifying complex human intents in practical situations. It is essential to use complementary knowledge of different modalities to improve the performance of intent comprehension.

Multimodal language understanding has attracted much attention in recent years. A series of multimodal datasets have been proposed in many areas such as sentiment analysis~\cite{MOSI, MOSEI, yu-etal-2020-ch}, humor detection~\cite{hasan2021humor}, sarcasm detection~\cite{castro-etal-2019-towards}, semantic comprehension~\cite{wang-etal-2019-youmakeup,yagcioglu-etal-2018-recipeqa}, etc. These benchmark datasets have extensively promoted the research and application of multimodal methodologies in related fields. However, there is still a lack of multimodal datasets for intent analysis. Most of the existing intent benchmark datasets contain merely the text modality~\citep{coucke2018snips, larson2019evaluation, liu2019benchmarking, casanueva2020efficient} or the visual modality~\cite{jia2021intentonomy}.  MDID~\cite{kruk-etal-2019-integrating} used image-caption pairs from Instagram posts to analyze multimodal intents, but the caption-based text information is different from spoken languages in the real world. 
\begin{table}[!htb]\small
	\caption{\label{statistics}  
		Statistics of the MIntRec dataset.}
	\begin{tabular}{ll}
		\toprule
		Total number of coarse-grained intents & 2\\
		Total number of fine-grained intents & 20\\
		Total number of videos  & 43\\
		Total number of video segments  & 2,224\\
		Total number of words in text utterances & 15,658\\
		Total number of unique words in  text utterances & 2,562\\
		Average length of text utterances & 7.04\\
		Maximum length of text utterances & 26\\
		Average length of video segments (s) & 2.38\\
		Maximum length of video segments (s) & 9.59\\
		\bottomrule
	\end{tabular}
\end{table}

The scarcity of data has seriously restricted the development of multimodal intent recognition. Nevertheless, constructing such a multimodal intent benchmark dataset faces two main challenges. Firstly, we need to design appropriate multimodal intent categories. Current intent taxonomies are mainly based on the text information or image-caption pairs, which have limitations when applied in multimodal scenes. Secondly, it requires distinguishing the visual information of the speaker as there is usually more than one person in the same situation. However, it will take much cost to perform manual annotation.  

To solve these problems, we propose a novel dataset, MIntRec, to fill the gap in multimodal intent recognition. The process of building the dataset is shown in Figure~\ref{pipeline}. Firstly, we prepare the original multimodal data for the dataset. The TV series SuperStore is selected as the data source due to its superiority for this task. After collecting the raw videos and subtitles, we process them into text utterances with respective video and audio segments. Then, we design both coarse-grained and fine-grained intent taxonomies for the multimodal scene. The coarse-grained taxonomies contain "Express emotions or attitudes" and "Achieve goals", which are inspired by the human intention philosophy~\cite{bratman1988intention}. Eleven and nine fine-grained intents are respectively summarized for these two coarse-grained categories based on the video segments and high-frequency intent tags.

\begin{table*}[t!]\small
	\caption{\label{guidelines} Intent taxonomies of our MIntRec dataset with brief interpretations.}
	\begin{tabular}{@{\extracolsep{0.001pt}}cll}
		\toprule
		\multicolumn{2}{c}{\textbf{Intent Categories}}&  
		\multicolumn{1}{c}{\textbf{Interpretations}} \\
		\midrule
		\multirow{15}{*}[0.5ex]{\begin{tabular}[l]{@{}c@{}}
				Express
				\\ 
				emotions
				\\
				or
				\\
				attitudes
		\end{tabular}}
		& Complain & \begin{tabular}[l]{@{}c@{}} Express dissatisfaction with someone or something  (e.g., saying unfair encounters with a sad expression and helpless motion). \end{tabular}  \\ 
		\cline{2-3}  \addlinespace[0.1cm]
		& Praise &  \begin{tabular}[l]{@{}l@{}} Express admiration for someone or something (e.g., saying with an appreciative  expression). \end{tabular}\\
		\cline{2-3}  \addlinespace[0.1cm]
		& Apologise & \begin{tabular}[l]{@{}l@{}} Express regret for doing something wrong (e.g., saying words of apology such as  sorry). \end{tabular}\\
		\cline{2-3}  \addlinespace[0.1cm]
		& Thank & \begin{tabular}[l]{@{}l@{}}Express gratitude in word or deed for the convenience or kindness given or offered by  others (e.g., saying  words\\ of appreciation such as thank you).\end{tabular}\\
		\cline{2-3}  \addlinespace[0.1cm]
		& Criticize & Point out someone's mistakes harshly (e.g., yelling out someone's problems).\\
		\cline{2-3}  \addlinespace[0.1cm]
		& Care &   \begin{tabular}[l]{@{}l@{}} Concern about someone or be curious about something (e.g., worrying about someone's health). \end{tabular}\\
		\cline{2-3}  \addlinespace[0.1cm]
		& Agree & \begin{tabular}[l]{@{}l@{}}Have the same attitude about something (e.g., saying affirmative words such as yeah and yes).\end{tabular}\\
		\cline{2-3}  \addlinespace[0.1cm]
		& Taunt & \begin{tabular}[l]{@{}l@{}} Use metaphors and exaggerations to accuse and ridicule (e.g., complimenting someone   with a negative expression).\end{tabular} \\
		\cline{2-3}  \addlinespace[0.1cm]
		& Flaunt & \begin{tabular}[l]{@{}l@{}} Boast about oneself to gain admiration, envy, or praise (e.g., saying something complimentary about oneself arrogantly).\end{tabular}\\
		\cline{2-3}  \addlinespace[0.1cm]
		& Oppose & Have an inconsistent attitude about something (e.g., saying negative words to express disagreement).\\
		\cline{2-3}  \addlinespace[0.1cm]
		& Joke & Say something to provoke laughter (e.g., saying something funny and exaggerated with a cheerful expression).\\
		\midrule
		\midrule
		%%%%%%%%%%%%%%%%%%%%%%%%%%
		\multirow{12}{*}[0.5ex]{\begin{tabular}[l]{@{}c@{}}
				Achieve
				\\
				goals
		\end{tabular}}
		& Inform & \begin{tabular}[l]{@{}l@{}}  Tell someone to make them aware of something (e.g., broadcasting something with a microphone). \end{tabular} \\
		\cline{2-3}  \addlinespace[0.1cm]
		& Advise &  \begin{tabular}[l]{@{}l@{}} Offer suggestions for consideration (e.g., saying words that make suggestions).\end{tabular}\\
		\cline{2-3}  \addlinespace[0.1cm]
		& Arrange & Plan or organize something (e.g., requesting someone what they should do formally).\\
		\cline{2-3}  \addlinespace[0.1cm]
		& Introduce & \begin{tabular}[l]{@{}l@{}} Communicate to make someone acquaintance with another or recommend something (e.g.,  describing a person's  identity \\ or the properties of an object). \end{tabular}\\
		\cline{2-3}  \addlinespace[0.1cm]
		& Comfort & \begin{tabular}[l]{@{}l@{}} Alleviate pain with encouragement or compassion (e.g., describing something is hopeful).  \end{tabular}\\
		\cline{2-3}  \addlinespace[0.1cm]		
		& Leave & Get away from somewhere (e.g., saying where to go while turning around or getting up).\\
		\cline{2-3}  \addlinespace[0.1cm]
		& Prevent & \begin{tabular}[l]{@{}l@{}} Make someone unable to do something (e.g., stopping someone from doing something with a hand). \end{tabular} \\
		\cline{2-3}  \addlinespace[0.1cm]
		& Greet & \begin{tabular}[l]{@{}l@{}}Express mutual kindness or recognition during the encounter (e.g., waving to someone and saying hello).\end{tabular}\\
		\cline{2-3}  \addlinespace[0.1cm]
		& Ask for help & Request someone to help (e.g., asking someone to deal with the trouble).\\ 
		\bottomrule
	\end{tabular}
\end{table*}
Next, we perform multimodal intent annotation with the prepared data and intent taxonomies. Five well-trained workers are employed for the annotation task. They label each sample among twenty intent tags with a convenient annotation platform, and the majority voting determines the multimodal labels. Finally, we obtain 2,224 high-quality samples for MIntRec. Besides, we propose an automatic process for speaker annotation. The detected object boundings are used to get the visual information of persons in each video frame. To identify the bounding boxes of speakers, we first detect and track faces within bounding boxes in different visual scenes and then predict the indexes of speakers with the active speaker detection algorithm. This process achieves high performance on our constructed testing set. 

After extracting features for each modality, we build baselines with three strong multimodal fusion methods. The experimental results show that leveraging the nonverbal information achieves $1\%\sim2\%$ stable improvements on both binary and multi-class classification. However, the results of the best methods are still far from human performance, indicating the challenge of the multimodal intent recognition task. 

Our contributions are summarized as follows:

(1) In this work, we build a novel multimodal intent recognition dataset, MIntRec, containing 2,224 high-quality samples with multimodal intent annotations. To the best of our knowledge, it is the first benchmark dataset for intent recognition in real-world multimodal scenes. 

(2) New intent taxonomies are designed for this task. Concretely, we provide two coarse-grained and twenty fine-grained intent categories for the study of multimodal intent analysis.

(3) An automatic speaker annotation process is proposed to produce high-quality annotated bounding boxes for speakers under the evaluation of over 12$\mathrm{K}$ human-annotated keyframes. It saves much time and laboratory and may benefit similar annotation tasks.

(4) Extensive experiments conducted on our dataset show utilizing multimodal information is superior to text-based intent recognition. The best-performing methods still have much room for improvement compared with humans. 

\section{MIntRec Dataset}
In this section, we will introduce the process of building the MIntRec dataset, including data preparation, intent taxonomy definition, multimodal intent annotation, and automatic speaker annotation. The detailed statistics of MIntRec are shown in Table~\ref{statistics}. 

\subsection{Data Preparation}
\label{data_collection}
Multimodal intent recognition requires plenty of nonverbal signals in real-world conversational scenes. For this purpose, we select the TV series Superstore as the source of our dataset, which has two main advantages: On the one hand, it consists of a wealth of characters (including seven prominent and twenty recurring roles) with different identities in the superstore, which is helpful to produce rich body language, expressions, and tones as multimodal information. On the other hand, it contains a mass of stories in various scenes (e.g., shopping mall, warehouse, office), which favor collecting diverse intent categories.

The raw videos and subtitles of Superstore are accessible on YouTube and OpenSubtitles\footnote{\url{https://www.opensubtitles.org/}}. To obtain video segments, we first extract each utterance's starting and ending timestamps of a speaker and then split the raw videos according to these timestamps. The corresponding audio segments are extracted from raw videos with the moviepy toolkit\footnote{https://pypi.org/project/moviepy/}. 

\subsection{Intent Taxonomy Definition}
Existing intent taxonomies are restricted in specific tasks~\cite{coucke2018snips,larson2019evaluation,casanueva2020efficient} or from social media posts~\cite{kruk-etal-2019-integrating,jia2021intentonomy}, which are uncommon in real-world scenes. Therefore, we design new taxonomies for multimodal intent recognition, including two coarse-grained and twenty fine-grained intent categories. 

In artificial intelligence research, intentions are regarded as plans or goals of an agent, accompanied by the corresponding feedback actions~\cite{bratman1988intention,wooldridge1995intelligent}. However, Schr{\"o}der~\cite{schroder2014intention} pointed out that the brain's emotional evaluations of situations are also critical components of human intentions. We combine these two aspects and crawl through the raw videos to generalize two representative coarse-grained intent taxonomies for multimodal intent recognition, including "Express emotions or attitudes" and "Achieve goals".

The coarse-grained intent taxonomy is insufficient to distinguish the complex and diverse types of human intentions in the real world. Thus, we further classify it into fine-grained categories. Firstly, we analyze different video segments as many as possible and collect several rough intent tags as candidates for each coarse-grained category. Then, we discuss and divide similar tags into the same group (e.g., introducing something or someone, worrying about someone or being interested in something). Next, we collect high-frequency intent tags and organize them into twenty fine-grained categories, including eleven classes for "Express emotions or attitudes" and nine for "Achieve goals". Some intents may be cued by a single modality such as text (e.g., thank, apologise, greet, agree, praise), video (e.g., leave, prevent), or audio (e.g., complain, criticize). Other intents may be inferred by combining different modalities (e.g., comfort, care, joke, taunt, flaunt). Brief interpretations of each intent category are summarized by observing practical examples and referring to related materials~\cite{dictionary1989oxford}, as shown in Table~\ref{guidelines}.
\begin{figure}[t!]\small
	\centering  
	\includegraphics[scale=.38]{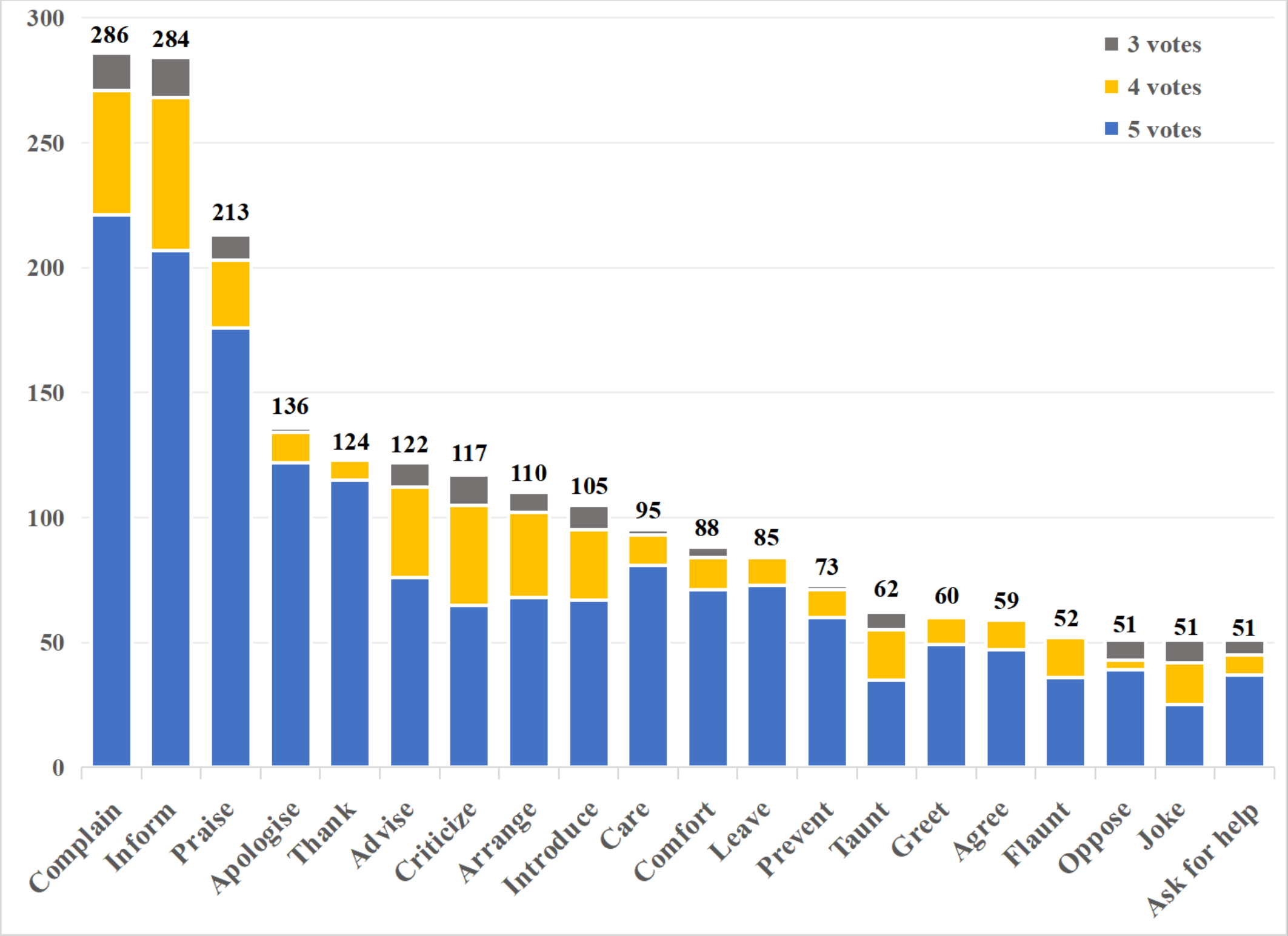}
	\caption{\label{detailed_votes} Voting statistics of 2,224 samples in MIntRec.
	}
\end{figure}
\subsection{Multimodal Intent Annotation}
After preparing data and defining intent taxonomies, we employ five students with an English foundation for annotation. Employees are offered interpretations and typical examples of each intent category as guidelines. Only well-trained employees are allowed for annotation. As intents usually exist in specific scenes of events~\cite{schroder2014intention}, there are irrelevant utterances among consecutive video segments, so we add a UNK tag to the label set to distinguish them. To improve the labeling efficiency,  we build a database to manage all the multimodal data and a convenient  platform for annotation. Users only need to click the button of the intent tag to complete annotation for a piece of data.

Each of the five workers is required to complete the annotation task of the same set of data independently. They need to choose the most confident intent tag for each sample by combining the video, audio, and text information.  The intent labels are determined by the majority voting (three out of five).  The samples with votes larger or equal to three (not UNK)
are saved. Finally, we acquired 2,224 high-quality samples to make up the MIntRec dataset. 

The detailed voting statistics are shown in Figure~\ref{detailed_votes}. It can be observed that intent categories with clear text or video cues (e.g., apologise, thank, leave, prevent, greet, agree) are easier to reach a higher agreement with votes larger than three. We also notice that the dataset is imbalanced. The reason is that it satisfies the distribution of different intents in real-world scenarios. Some intents occur more frequently (e.g., complain, inform, and praise), while others do not (e.g., joke and ask for help). Nevertheless, each intent category still contains at least fifty samples. 

\subsection{Automatic Speaker Annotation}
As suggested in~\cite{MOSI, MOSEI, yu-etal-2020-ch}, we first extract frames from each video segment to represent the video information. Then, we aim to annotate the visual contents related to the speakers, which are the objects of multimodal intent recognition. 

We perform object detection on the video frames to obtain rich visual information containing facial and body features. Specifically,  we use a Faster R-CNN~\cite{ren2015faster} with the backbone ResNet-50~\cite{he2016deep} pre-trained on the MS COCO dataset~\cite{lin2014microsoft} (containing 250,000 person instances with seventeen keypoints) to predict the bounding boxes of persons in each frame. 

However, there are still two challenges for speaker annotation. For one thing, there may be no or little visual information about the speaker in the extracted frame (e.g., most of the body is covered, or the speaker only appears on the back of the body). For another thing, there are usually multiple persons with detected bounding boxes in each frame, making it hard to distinguish the speaker. To solve these problems, we propose an automatic process to perform speaker annotation with the aid of the audio-visual active speaker detection algorithm~\cite{chung2016out,tao2021someone}. It contains the following four steps:

Firstly, we use the scene detection toolkit\footnote{\url{https://pypi.org/project/scenedetect/}} to distinguish different visual scenes in a video segment, as there may be a change in contents between adjacent frames. Secondly, a pre-trained Faster R-CNN is used to detect the bounding boxes of persons for each frame in a visual scene. Since facial motion (e.g., lips movement) is critical for detecting speakers, we further use the S$^{3}$FD~\cite{zhang2017s3fd} algorithm to detect faces in the bounding boxes and establish the one-to-one mapping between faces and object boundings. Note that this step also filters the keyframes with clear facial features. Thirdly, a simple and effective method is used to perform face tracking. Concretely, we compute intersection-over-union (IoU) for two faces in adjacent frames and consider two faces are from the same person if IoU is above 0.5. Given that accidental objects may block faces, we tolerate up to ten consecutive frames of missing faces. Finally, we use a pre-trained TalkNet model with tracked faces and corresponding audio information to predict speakers and determine respective bounding boxes with the mapping obtained in the second step. With the aid of this process, we automatically generate more than 120$\mathrm{K}$ keyframes with speaker annotations of bounding boxes free from any manual intervention. 

To evaluate the quality of keyframes and bounding box information, we construct a testing set with more than 12$\mathrm{K}$ human-annotated keyframes. Specifically, we first uniformly extract one shot every ten frames and manually select keyframes with clear visual information. Then, a pre-trained Faster R-CNN is used to predict the object boundings of persons in each keyframe, and annotators label the speakers by choosing the indexes of corresponding bounding boxes.

Compared with the human-annotated keyframes, the missing rate of generated keyframes is only 2.3$\%$. Among the hit keyframes, the proportion of high-quality predicted bounding boxes  (IoU $>$ 0.9) is 90.9$\%$. The evaluation results demonstrate the reliability of the automatic speaker annotation process. Besides, it is much more efficient without labor-intensive and time-consuming manual annotations. 

\section{Methodology}
After preparing the corresponding text, video, and audio data of speakers, we extract features of each modality and use them for multimodal fusion.

\subsection{Feature Extraction}

\subsubsection{Text} Due to the excellent performance of the pre-trained BERT language model in the Natural Language Processing (NLP) community~\cite{BERT}, we use it to extract text features. For each text utterance, we obtain the token embeddings   $\mathbf{z}^{T} \in \mathds R^{L_{T} \times H_{T}}$ from the last hidden layer of BERT, where $L_{T}$ is the sequence length of text utterances, and $H_{T}$ is the feature dimension 768.

\subsubsection{Vision} The object detection method is used for extracting vision features. For each video segment, we first leverage a pre-trained Faster R-CNN with the backbone ResNet-50 to extract the feature representations $\mathbf{x}$ of all keyframes. Then, we map $\mathbf{x}$ into the regions with the annotated bounding boxes $B$ to obtain the vision feature embeddings $\mathbf{z}^{V} \in \mathds R^{L_{V} \times H_{V}}$:
\begin{align}
\mathbf{z}^{V} = \text{AvgPool} (\text{RoIAlign} (\mathbf{x}, B)),
\end{align}
where RoIAlign~\cite{he2017mask} is used to extract the fixed size feature maps (e.g., 7$\times$7). \text{AvgPool} is used to reduce both weight and height sizes to the unit size. $L_{V}$ is the sequence length of video segments, and $H_{V}$ is the feature dimension 256.

\subsubsection{Audio}
The speech toolkit librosa~\cite{mcfee2015librosa}  is first used to acquire audio time series at 16,000 Hz. Then, the pre-trained model wav2vec 2.0~\cite{baevski2020wav2vec}   is used to extract audio features, which learns powerful representations for speech recognition with self-supervised learning. We obtain the acoustic feature embeddings $\mathbf{z}^{A} \in \mathds R^{L_{A} \times H_{A}}$ from the last hidden layer of wav2vec 2.0, where $L_{A}$ is the sequence length of audio segments, and $H_{A}$ is the feature dimension 768.

\subsection{Benchmark Multimodal Fusion Methods}
\label{benchmark}
After feature extraction, we benchmark three powerful multimodal fusion methods for the MIntRec dataset. These methods aim to learn the interactions between different modalities with the extracted features and obtain friendly representations for multimodal fusion. 

\subsubsection{MulT}
The Multimodal Transformer (MulT)~\cite{tsai2019multimodal} is an end-to-end method to deal with non-aligned multimodal sequences. It extends the vanilla Transformer~\cite{Transformer} to the cross-modal Transformer with the pairwise inter-modal attention mechanism, which helps to capture the adaptation knowledge between different modalities in the latent space.

\subsubsection{MISA}
Hazarika et al.~\cite{hazarika2020misa} proposed the framework MISA to learn multimodal representations with modality-invariant and modality-specific properties. On the one hand, a shared subspace is utilized to learn common features of all modalities. On the other hand, distinct subspaces are designed to capture the unique attributes of each modality. For this purpose, the training objectives contain four aspects: similarity loss, difference loss,  reconstruction loss, and task-specific loss.

\subsubsection{MAG-BERT}
Rahman et al.~\cite{BERT_MAG} integrated two nonverbal modalities into BERT with an additional multimodal adaptation gate (MAG) module. MAG can produce a position shift in the semantic space adaptive to acoustic and visual information. It can be flexibly placed between layers of BERT to receive inputs from nonverbal modalities. 

In this work, the features of each modality $\mathbf{z}^{T}$,$ \mathbf{z}^{V}$, and $\mathbf{z}^{A}$ can be directly used as the inputs of MulT and MISA. As MAG-BERT needs aligned multimodal data, we pass the features of video and audio ($ \mathbf{z}^{V}$ and  $\mathbf{z}^{A}$) through the Connectionist Temporal
Classification (CTC)~\cite{graves2006connectionist} module to align with the text feature $\mathbf{z}^{T}$ in the word-level as suggested in~\cite{tsai2019multimodal}.  For each method, we use the multimodal annotations as targets and perform the classification task under the supervision of the softmax loss.
\begin{table}[t!]\small
	\caption{\label{datasplit} Dataset splits in MIntRec. The training, validation and testing sets are split into 3:1:1. 
	}
	\begin{tabular}{l|c|cccc} 
		\midrule
		Item  & Total  & Express emotions or attitudes   &  Achieve goals \\
		\midrule
		Train  & 1,334   & 749  & 585    \\
		Valid  & 445  & 249 & 196  \\
		Test   & 445   & 248  & 197 \\
		\midrule
	\end{tabular}
\end{table}
\begin{table*}[t!]\small
			\caption{\label{results-main-1}  
			Multimodal intent recognition results on the MIntRec dataset. "Twenty-class" and "Binary" denote the multi-class and binary classification over fine-grained and coarse-grained intent taxonomies. $\Delta$ denotes the most improvement over the text-classifier baseline in the current evaluation metric of each method.}
	\centering
	\begin{tabular}{@{\extracolsep{3pt}}l|c|cccc|cccc}
		\toprule
		\multirow{2}{*}{Methods} & \multirow{2}{*}{Modalities} 
		& \multicolumn{4}{c|}{Twenty-class}& \multicolumn{4}{c}{Binary}\\
		
		&& ACC  & F1  & P  & R & ACC  & F1  & P & R \\ 
		\midrule
		Classifier & Text & 70.88 & 67.40 & 68.07 & 67.44 & 88.09 & 87.96 &	87.95 &	88.09\\
		\midrule
		\multirow{4}{*}{MAG-BERT} & Text + Audio & 72.16 & 68.28 & 68.88 &	68.88 &	88.83 &	88.71 &	88.67 &	88.85  \\
		& Text + Video & 72.09 & 67.92 & \textbf{69.09} & 68.73 &	88.45 &	88.28 &	88.36 &	88.27 \\
		& Text + Audio + Video & \textbf{72.65} & \textbf{68.64} & 69.08 & \textbf{69.28}  & \textbf{89.24} & \textbf{89.10} & \textbf{89.10}  & \textbf{89.13} \\ 
		& $\Delta$ & 1.77$\uparrow$ & 1.24$\uparrow$ & 1.02$\uparrow$ & 1.84$\uparrow$  & 1.15$\uparrow$ & 1.14$\uparrow$ & 1.15$\uparrow$  & 1.04$\uparrow$ \\ 
		
		\midrule
		\multirow{4}{*}{MulT} & Text + Audio & 71.80 &	67.95 &	69.18 &	67.96 &	88.74 &	88.61 &	88.59 &	88.68 \\
		& Text + Video & 71.98 & 68.76 & 69.68 & 68.79 & 88.79 & 88.66 & 88.63 & 88.77\\
		& Text + Audio + Video & \textbf{72.52} & \textbf{69.25} & \textbf{70.25} & \textbf{69.24}  & \textbf{89.19} & \textbf{89.07} & \textbf{89.02}  & \textbf{89.18} \\
		& $\Delta$& 1.64$\uparrow$ & 1.85$\uparrow$ & 2.18$\uparrow$ & 1.80$\uparrow$  & 1.10$\uparrow$ & 1.11$\uparrow$ & 1.07$\uparrow$  & 1.09$\uparrow$ \\ 
		
		\midrule
		\multirow{4}{*}{MISA}& Text + Audio & 71.60 & 68.37 & 69.57 &	68.30 &	88.45 &	88.31 &	88.32 &	88.35  \\
		& Text + Video & 71.53 & 68.34 & 69.68 & 68.19 & 88.74 &	88.60 &	88.63 &	88.65 \\
		& Text + Audio + Video & \textbf{72.29} & \textbf{69.32} & \textbf{70.85} & \textbf{69.24}  & \textbf{89.21} & \textbf{89.06} & \textbf{89.12}  & \textbf{89.06} \\ 
		& $\Delta$ & 1.41$\uparrow$ & 1.92$\uparrow$ & 2.78$\uparrow$ & 1.80$\uparrow$  & 1.12$\uparrow$ & 1.10$\uparrow$ & 1.17$\uparrow$  & 0.97$\uparrow$ \\ 
		\midrule
		\multirow{2}{*}{Human} 
		& - & 
		85.51 & 
		85.07 & 
		86.37 & 
		85.74 & 
		94.72 & 
		94.67 & 
		94.74 & 
		94.82 \\
		& $\Delta$ & 14.63$\uparrow$ & 17.67$\uparrow$ & 18.30$\uparrow$ &
		18.30$\uparrow$ & 6.63$\uparrow$ & 6.71$\uparrow$ & 6.79$\uparrow$  & 6.73$\uparrow$ \\
		\bottomrule 
	\end{tabular}
	%modality, methods, standard deviations / mean, 10
\end{table*}  
\section{Experiments}
This section introduces the experimental setup, baselines, and experimental results. 

\subsection{Experimental Setup}
\subsubsection{Dataset Splits}
We shuffle the video segments in random and split training, validation, and testing sets by multimodal annotations in 3:1:1. The detailed statistics are shown in Table~\ref{datasplit}. 

\subsubsection{Evaluation Metrics}
Four metrics are used to evaluate the model performance:
accuracy (ACC), F1-score (F1), precision (P), and recall (R). We report the macro score over all classes for the last three metrics. The higher values indicate better performance of all metrics. 

\subsubsection{Implementation Details}
For the text and audio modalities, we employ the pre-trained BERT model (bert-base-uncased, with 12 Transformer layers) and pre-trained wav2vec 2.0 model implemented in PyTorch~\cite{wolf2020transformers}. For the video modality, we use a pre-trained Faster R-CNN with  ResNet-50 backbone implemented in  MMDetection Toolbox~\cite{chen2019mmdetection}.

As sequence lengths of the segments in each modality need to be fixed, we use zero-padding for shorter sequences. $L_{T}$, $L_{V}$, and $L_{A}$  are 30, 230, and 480, respectively. For all methods, the training batch size is 16, and the number of training epochs is 100. We adjust the hyper-parameters with macro F1-score. For a fair comparison, we report the average performance over ten runs of experiments with random seeds 0-9. 

\subsection{Baselines}
We build a series of baselines for the MIntRec dataset. As the text modality is predominant in the intent recognition task, we train a classifier with the text-only modality as the primary baseline. As suggested in~\cite{BERT}, we use the first special token [CLS] from the last hidden layer as the sentence representation and fine-tune the pre-trained BERT model with the downstream classification task for better performance.  

As introduced in section~\ref{benchmark}, three multimodal fusion methods, MAG-BERT, MulT, and MISA, are used to benchmark our dataset. Besides, we also modify them to use two modalities (Text + Audio and Text + Video) as inputs for ablation studies. 

We have a different set of two annotators to evaluate the human performance on this task. They are provided with the training and validation sets with multimodal annotations for learning and assessment as in baselines. After that, they need to label the unseen testing set, and their average results are reported. 

\subsection{Results}
We conduct experiments on the MIntRec dataset with several baselines and show the results in Table~\ref{results-main-1}. For each multimodal fusion method, the best results are highlighted in bold. The improvements over the text-classifier are shown with $\Delta$.

The multimodal fusion methods achieve substantial improvements on all metrics of twenty-class and binary classification compared with the text-only modality. All the multimodal fusion methods for twenty-class classification stably improve over 1\% scores on all metrics. All the baselines achieve much higher performance on binary classification. We suggest the reason is that recognizing coarse-grained intent categories is much easier than distinguishing fine-grained intent categories. Nevertheless, all the multimodal fusion methods still yield over 1\% improvements on almost all metrics. The results demonstrate that effectively leveraging the multimodal information helps enhance the intent recognition capability. 
\begin{table*}[t!]\small
	\caption{\label{emotion_results_each_class} Results of each fine-grained intent category in "Express emotions and attitudes".}
	\renewcommand\tabcolsep{3pt} 
	\centering
	
	\begin{tabular*}{\textwidth}{l|c@{\extracolsep{\fill}}lccccccccccc} 
		\toprule
		% \multirow{2}{*}{Methods} & \multicolumn{11}{c}{Intent Classes} \\
		Methods &Complain& Praise & Apologise & Thank & Criticize  & Care & Agree & Taunt & Flaunt & Oppose & Joke   \\ 
		\midrule
		\multicolumn{1}{l|}{Text-classifier}  & 64.36 & 85.69 & 97.93 & 97.22 & 47.06 & 87.42 & 94.26 & 15.53 & 46.12 & 32.32 & 27.42 \\
		\midrule
		\multicolumn{1}{l|}{MAG-BERT}  & \textbf{67.65} & 86.03 & 97.76 & 96.52 & 49.02 & 85.59 & 91.60 & 15.78 & 47.09 & 33.97 & 37.54\\
		\multicolumn{1}{l|}{MulT} & 65.48 & 84.72 & \textbf{97.93} & 96.83 & 49.72 & \textbf{88.12} & \textbf{92.23} & \textbf{26.12} & \textbf{48.91} & 34.68 & 33.95 \\
		\multicolumn{1}{l|}{MISA} & 63.91 & \textbf{86.63} & 97.78 & \textbf{98.03} & \textbf{53.44} & 87.14 & 92.05 & 22.15 & 46.44 & \textbf{36.15} & \textbf{38.74}\\
		\multicolumn{1}{l|}{$\Delta$} &3.29$\uparrow$ & 0.94$\uparrow$ & 0.00 & 0.81$\uparrow$ & 6.38$\uparrow$ & 0.70$\uparrow$ & 2.03$\downarrow$ & 10.59$\uparrow$ & 2.79$\uparrow$ & 3.83$\uparrow$ & 11.32$\uparrow$ \\
        \midrule
		\multicolumn{1}{l|}{Human}  & 80.08 & 93.44 & 96.15 & 96.90 & 72.21 & 96.09 & 87.21 & 65.55 & 78.10 & 69.04 & 72.22 \\
		\multicolumn{1}{l|}{$\Delta$}  & 15.72$\uparrow$ & 7.75$\uparrow$ & 1.78$\downarrow$ & 0.32$\downarrow$ & 25.15$\uparrow$ & 8.67$\uparrow$ & 7.05$\downarrow$ & 50.02$\uparrow$ & 31.98$\uparrow$ & 36.72$\uparrow$ & 44.80$\uparrow$ \\
		\midrule
	\end{tabular*}
\end{table*}

\begin{table*}\small
		\caption{\label{goal_results_each_class} Results of each fine-grained intent category in "Achieve goals".
	}
	\renewcommand\tabcolsep{3pt} 
	\centering
		\begin{tabular*}{\textwidth}{l|c@{\extracolsep{\fill}}lccccccccccc} 
		\midrule
		Methods & Inform  &Advise & Arrange & Introduce & Comfort & Leave & Prevent & Greet & Ask for help\\
		\midrule
		\multicolumn{1}{l|}{Text-classifier}  & 67.74 & 67.68 & 64.67 & 68.64 & 77.05 & 73.37 & 82.47 & 84.90 & 66.20   \\
		\midrule
		\multicolumn{1}{l|}{MAG-BERT}  & \textbf{71.00} & 69.30 & 63.82 & 67.42 & 76.43 & 75.77 & \textbf{85.07} & \textbf{91.06} & 64.44 \\
		\multicolumn{1}{l|}{MulT}  & 
        70.85 & 69.43 & 65.44 & \textbf{71.19} & 76.44 & 75.58 & 81.68 & 86.65 & \textbf{69.12} \\
		\multicolumn{1}{l|}{MISA} & 
		70.18 & \textbf{69.56} & \textbf{67.32} & 67.22 & \textbf{78.78} & \textbf{77.23} & 83.30 & 82.71 & 67.57   \\
		\multicolumn{1}{l|}{$\Delta$} & 3.26$\uparrow$ & 1.88$\uparrow$ & 2.65$\uparrow$ & 2.55$\uparrow$ & 1.73$\uparrow$ & 3.86$\uparrow$ & 2.60$\uparrow$ & 6.16$\uparrow$ & 2.92$\uparrow$ \\
		\midrule
		\multicolumn{1}{l|}{Human}  & 79.69 & 87.14 & 81.40 & 84.09 & 95.95 & 97.06 & 86.43 & 94.15 & 88.54 \\
		\multicolumn{1}{l|}{$\Delta$}  & 11.95$\uparrow$ & 19.46$\uparrow$ & 16.73$\uparrow$ & 15.45$\uparrow$ & 18.90$\uparrow$ & 23.69$\uparrow$ & 3.96$\uparrow$ & 9.25$\uparrow$ & 22.34$\uparrow$ \\
		\bottomrule
	\end{tabular*}
\end{table*}

However, even the best-performing methods are still far away from humans. Compared with the text modality, the human performance improves by $14\%\sim19\%$ on twenty-class classification and $6\%\sim7\%$ on binary classification. The improvements are much more significant than in multimodal fusion methods, indicating this task is very challenging for multimodal research.

\section{Discussion}
This section analyzes the effect of nonverbal modalities and shows the performance of fine-grained intent categories with quantitative results.
\subsection{Effect of Nonverbal Modalities}
We conduct ablation studies for each multimodal fusion method to investigate the influence of the video and audio modalities. Specifically, we compare the tri-modality with bi-modality and show results in Table~\ref{results-main-1}.

\subsubsection{Bi-modality} 
After combining text with audio modality, the intent recognition performance achieves overall gains on both twenty-class and binary classification. Specifically, MAG-BERT, MulT, and MISA  increase accuracy scores of 1.28\%, 0.92\%, and 0.72\% on twenty-class and 0.74\%, 0.65\%, and 0.36\% on binary classification, respectively. Combining text with video modality also leads to better performance in all settings. Similarly, MAG-BERT, MulT, and MISA increase accuracy scores of 1.21\%, 1.10\%, and 0.65\% on twenty-class and 0.36\%, 0.70\%, and 0.65\% on binary classification. 

Due to the consistent improvements in leveraging video or audio modality, we suppose the two nonverbal modalities are critical for multimodal intent recognition. The valuable information such as tone of voice and body movements may be helpful to recognize human intents from new dimensions. 

\subsubsection{Tri-modality}
Though multimodal fusion methods with bi-modality have achieved better performance than the text modality, we find utilizing the tri-modality brings more gains. MAG-BERT achieves a slight advantage on the precision score but performs worse on the other metrics. The positive results demonstrate that both video and audio modalities benefit this task. The benchmark multimodal fusion methods can fully use the information from different modalities by modeling cross-modal interactions.

\subsection{Performance of Fine-grained Intent Classes}
To investigate the effect of the multimodal information in each fine-grained intent category, we report the average macro F1-score of each class over ten runs of experiments for all baselines and show results in Table~\ref{emotion_results_each_class} and~\ref{goal_results_each_class}. The best results of multimodal fusion methods are highlighted in bold. $\Delta$ indicates the most improvement of multimodal fusion methods and humans over the text-classifier. 

Firstly, we observe the results of each class in the coarse-grained intent category "Express emotions and attitudes" in Table~\ref{emotion_results_each_class}. It is shown that multimodal fusion methods perform better than text-classifier in most classes. Notably, we find there are some classes with over $3\%$ significant improvements (e.g., complain, criticize, taunt, oppose, joke). The success of these intents is intuitive, as they contain vivid nonverbal signals of expressions and tones, requiring the aid of visual and audio information. However, we also notice that the multimodal information is less helpful with few improvements or even degradation in some classes (e.g., apologise, thank, praise, agree). The reason is that these classes usually contain clear textual cues such as sorry, thank, yeah, etc. In this case, the pre-trained language model is good enough for intent recognition. 

% Some classes achieve excellent performance with the text modality alone (e.g., apologise, thank, agree) as they have clear textual 
% cues for intent recognition such as sorry, thank, yeah, etc. The multimodal information is less helpful in some difficult classes (e.g., care, flaunt), as they are easily confused with other categories.

Secondly, we observe the results of each class in the coarse-grained intent category "Achieve goals" in Table~\ref{goal_results_each_class}. The performance of multimodal fusion methods consistently achieves over 1$\%\sim6\%$ improvements in all classes. It is reasonable because these classes are highly associated with broad body movements, such as gestures, posture, arm behaviors, etc. By comparison, capturing this information with the text modality is rather challenging. 

Finally, we observe the human performance in Table~\ref{emotion_results_each_class} and~\ref{goal_results_each_class}. As expected, humans have gained an absolute advantage over the text modality in most intent categories. However, the human performance is lower than text-classifier in three classes (apologise, thank, agree). It suggests that even humans may make mistakes in the classes biased to the text modality. In contrast, humans are good at reasoning through different modalities and show significant superiority with over 10\% improvements in many intents, such as taunt, flaunt, oppose, joke, etc. The huge gap indicates the necessity of exploring an effective way to leverage nonverbal information.

\section{Related Works}

\subsection{Multimodal Language Datasets}
Multimodal language understanding is a booming area with a series of emerging benchmark datasets. For example, many datasets have been proposed in multimodal sentiment analysis~\cite{busso2008iemocap,MOSI,MOSEI,yu-etal-2020-ch} and emotion recognition~\cite{poria2019meld}. Some multimodal datasets also detect unique properties of human languages, such as sense of humor~\cite{ hasan2021humor, hasan2019ur}, metaphor~\cite{zhang-etal-2021-multimet}, sarcasm~\cite{castro-etal-2019-towards,cai-etal-2019-multi}. Moreover, multimodal datasets are designed for a series of other tasks in NLP, such as dialogue act classification~\cite{saha2020towards,saha2021emotion}, named entity recognition~\cite{sui-etal-2021-large}, comprehension and reasoning~\cite{yagcioglu-etal-2018-recipeqa,wang-etal-2019-youmakeup}, comments generation~\cite{wang2020videoic}, fake news detection~\cite{nakamura2020fakeddit}, etc. Nevertheless, there is a lack of multimodal datasets for intent analysis in real-world dialogue scenes.

\subsection{Benchmark Datasets for Intent Analysis}
Intent analysis is a popular research field in NLU, and many important tasks have been proposed, such as joint intent detection and slot filling~\cite{qin-etal-2019-stack,zhang2019joint}, open intent detection~\cite{lin-xu-2019-deep,Zhang_Xu_Lin_2021} and discovery~\cite{lin2020discovering,Zhang_Xu_Lin_Lyu_2021}. The booming of this area benefits from several benchmark intent datasets proposed in recent years, such as  ATIS~\cite{hemphill1990atis}, Snips~\cite{coucke2018snips}, CLINC150~\cite{larson2019evaluation}, HWU64~\cite{liu2019benchmarking}, and BANKING77~\cite{casanueva2020efficient}. These datasets collected the corpus by interacting with the intelligent assistant or customers in specific domains and used the crowdsourcing task among service requests to determine intent labels. StackOverflow~\cite{xu-etal-2015-short} and StackExchange~\cite{braun-etal-2017-evaluating} gathered data from technical question and answering platforms. Their intent labels are defined as the tags assigned to the questions. SWBD~\cite{godfrey1992switchboard} corpus contained 42 dialogue acts (DAs) for task-independent conversations. Still, many DAs are ambiguous concepts (e.g., statement-opinion and statement-non-opinion), which are difficult to be applied in real applications. Intentonomy~\cite{jia2021intentonomy} analyzed the visual intents among social media posts and collected an image dataset. However, all these datasets merely contain information from a single modality.

MDID~\cite{kruk-etal-2019-integrating} integrated image and text information for intent recognition. However, the multimodal information from Instagram posts is limited, and the taxonomies are inappropriate in real-world scenes. In contrast, MIntRec contains rich multimodal information in dialogue scenes with text, video, and audio modalities. 

\subsection{Multimodal Fusion Methods}
Based on the multimodal language datasets,  multimodal fusion methods are proposed to capture the interactions between language and nonverbal modalities. Traditional methods, such as
MCB~\cite{fukui2016multimodal} and TFN~\cite{zadeh2017tensor}  obtained representations by learning intra-modality and inter-modality relations. However, the high-dimensional representations suffer from high computational complexity. 
LMF~\cite{LMF} designed low-rank multimodal tensors to solve this problem. MFN~\cite{MFN} first learned view-specific interactions for every single modality and then used the attention mechanism to summarize cross-perspective interactions through the multi-perspective gated memory. 

Recent methods adopt Transformer-based methods for multimodal representation learning. For example, MulT~\cite{tsai2019multimodal} managed to learn interactions between different modalities with directional cross-modal attention. MISA~\cite{hazarika2020misa} performed multimodal fusion with multi-headed self-attention to capture the relations between modality-invariant and modality-specific representations. MAG-BERT~\cite{BERT_MAG} introduced the multimodal adaptation gate for pre-trained Transformers to receive information from different modalities. In this work, we adapt the above three algorithms to multimodal intent recognition as benchmark methods. 

\subsection{Audio-visual Active Speaker Detection}
Active speaker detection (ASD) aims to detect the speaker(s) in a visual scene. In this work, we focus on ASD with audio and visual information. 
Some studies~\cite{chung2016out, afouras2020self} treated this problem as a binary classification task and used a multi-layer perceptron (MLP) for ASD with concatenated audio and visual features. Besides, temporal structures~\cite{tao2019end,roth2020ava} such as recurrent neural networks (RNNs) were adopted to obtain better performance with time-series information.  MAAS~\cite{alcazar2021maas} introduced graph convolutional networks (GCNs) to model interactions between audio and video modalities. TalkNet~\cite{tao2021someone} introduced an audio-visual cross-attention mechanism for effectively modeling cross-modal interactions and a self-attention mechanism for capturing long-term speech dependencies. In this work, we use TalkNet for automatic speaker annotation. 

\section{Conclusions}
This paper first presents a new dataset for multimodal intent recognition. It has 2,224 high-quality annotated samples with corresponding multimodal information. New taxonomies with coarse-grained and fine-grained intent categories are specifically designed for real-world multimodal scenes. We also propose an automatic process to obtain the information of object boundings towards speakers, which vastly reduces the annotation burden. We make great efforts to ensure the quality of our dataset and build baselines with three multimodal fusion methods. Comprehensive experiments verify the superiority of multimodal information for intent recognition. The gap between the best-performing multimodal fusion methods and humans indicates there is still a long way to go for multimodal intent recognition.

\begin{acks}
This paper is founded by National Natural Science Foundation of China (Grant No. 62173195) and Beijing Academy of Artificial Intelligence (BAAI). We would like to thank Guohui Guan, Wenrui Li, and Xiaofei Chen for their efforts during dataset construction.
\end{acks}

\bibliographystyle{ACM-Reference-Format}
% \bibliography{MM}
\balance
\bibliography{sample-sigconf}

\end{document}